\newcommand{\CLASSINPUTtoptextmargin}{0.76in}
\newcommand{\CLASSINPUTbottomtextmargin}{1.1in}
\documentclass[conference]{IEEEtran}
\IEEEoverridecommandlockouts
\usepackage{cite}
\usepackage{amsmath,amssymb,amsfonts}

\usepackage{graphicx}
\usepackage{textcomp}
\usepackage{xcolor}
\usepackage{multirow}
\usepackage{makecell}
\usepackage{float}
\usepackage{capt-of}
\usepackage{ulem}
\usepackage{subfigure}
\usepackage{threeparttable}
\definecolor{sig}{RGB}{217,150,144}
\definecolor{reg}{RGB}{1,160,160}
\definecolor{exp}{RGB}{128, 0,255}
\definecolor{man}{RGB}{0,200,0}

\makeatletter
\newif\if@restonecol
\makeatother

\usepackage[linesnumbered,ruled]{algorithm2e}%[ruled,vlined]{
\usepackage{algpseudocode}
\def\BibTeX{{\rm B\kern-.05em{\sc i\kern-.025em b}\kern-.08em
    T\kern-.1667em\lower.7ex\hbox{E}\kern-.125emX}}
\begin{document}

\title{Training Deep Neural Networks Using Posit Number System
}

\author{
         Jinming~Lu,
         Siyuan~Lu,
         Zhisheng~Wang,
         Chao~Fang,	
 		 Jun~Lin,
         Zhongfeng~Wang,
         and Li~Du\\
 		\IEEEauthorblockA{School of Electronic Science and Engineering, Nanjing University, Nanjing, China\\	
 		Email: \{jmlu,~sylu,~zswang\}@smail.nju.edu.cn, fantasysee@foxmail.com, \{jlin,~zfwang\}@nju.edu.cn, dl1989113@ucla.edu}
}

\maketitle

\begin{abstract}
With the increasing size of Deep Neural Network (DNN) models, the high memory space requirements and computational complexity have become an obstacle for efficient DNN implementations. To ease this problem, using reduced-precision representations for DNN training and inference has attracted many interests from researchers. This paper first proposes a methodology for training DNNs with the posit arithmetic, a type-3 universal number (Unum) format that is similar to the floating point(FP) but has reduced precision. A warm-up training strategy and layer-wise scaling factors are adopted to stabilize training and fit the dynamic range of DNN parameters. With the proposed training methodology, we demonstrate the first successful training of DNN models on ImageNet image classification task in 16 bits posit with no accuracy loss.
Then, an efficient hardware architecture for the posit multiply-and-accumulate operation is also proposed, which can achieve significant improvement in energy efficiency than traditional floating-point implementations. The proposed design is helpful for future low-power DNN training accelerators.
\end{abstract}


\section{Introduction}
% 介绍 DNN， DNN compression ,DNN training compression
% DNN目前在许多任务中达到了STOA的结果，例如xxx. 随着网络的变得越来越复杂，对于计算量和存储的需求成为严重的问题
Recently deep neural networks (DNNs) have made a great success in many real-world applications, such as image classification \cite{he2016deep}, speech recognition \cite{amodei2016deep}, and natural language processing \cite{wang2016attention}.
With the increasing size of DNNs, the models show the state-of-the-art performance.
However, the high memory space requirements and computational complexity have become a serious problem for efficient implementations, especially on mobile devices.

% 为了解决这样的问题， 一些压缩方法呗提了出来。 比如xxx。
% 但是大多数针对inference阶段，对training的工作还有待探索。
% training 提出了更高的要求
To alleviate the extremely high demand of computational resource, many compression methods  are proposed, which aim to generate compact DNN models.
At present, the reduced-precision representation of numbers, also known as quantization, is one of the most attractive topics\cite{krishnamoorthi2018quantizing}.
However, these methods mainly focus on the inference phase of DNN.
Researches on training with limited-precision numbers still remain to be explored.

Because of the existence of more information flows, including gradients backpropagation and parameters updating,
the training of DNNs needs higher representation ability for data.
In other words, a suitable number format for DNN training should have enough dynamic range for big numbers,
and have high precision for numbers in the center of data distribution.

Posit, a type-3 universal number, is introduced by Gustafson \textit{et al.}\cite{gustafson2017beating}.
An $n$-bit posit number is defined as $(n,es)$, where $es$ (exponent bits) is used to control dynamic range.
Comparing to standard floating point(FP) number, posit has a better trade-off between dynamic range and precision,
just meeting the needs of low-bits number for DNN training.
% 有点毛病!!!!!!!!!
Some researchers have claimed the prospect of posit in DNNs, but practical implementations and verifications are absent\cite{gustafson2017beating}\cite{zhang2019efficient}.
% Motivation 不够强!!!!!!!!!!
In this paper, we first propose an effective strategy for DNN training using posit number system.
After the posit being proved useful in DNN training,
a processing element supporting posit arithmetics is required to make full use of its efficiency in DNN accelerators.
Our contributions are summarized as follows:

\begin{itemize}
    \item With an operation which transforms a real number to posit format, we illustrate how to apply the posit in DNN training process.
    % 还是想想怎么改 ~~
    \item We analyze the advantages and disadvantages of the application of posit in DNN training, then we propose corresponding solutions to overcome these problems.
    Firstly, to deal with the high sensitivity of models in the early training stage and ensure the convergence of models, a warm-up training with FP32 is carried out.
    Secondly, to take the advantage of posit, we design a layer-wise scaling factors based on the center of data distribution in log-domain,
    making the data distribution of models match the change of the precision of posit number.
    Thirdly, to meet different data ranges of different layers, we come up with a quanlitative criteria to select a proper $es$ to achieve a better trade-off between dynamic range and precision of posit number.
    \item
    In order to verify the effectiveness of our methods, ResNet-18 models are trained on ImageNet dataset and Cifar-10 dataset,
    where 8-bit or 16-bit posit numbers are applied in forward and backward computation, respectively.
    The experiments show no accuracy loss with the baseline model.
    \item
    We propose a hardware architecture for posit multiply-and-accumulate (MAC) unit, which is coded by Verilog HDL
    and synthesized by Design Compiler under TSMC 28nm technology.
    Comparing to standard floating point MAC unit, the posit MAC can reduce the power by 83\%,
    and reduce the area by 76\%.
    It demonstrates that our design will benefit future low-power DNN training accelerators.
\end{itemize}

% 用verilog 实现， 用啥综合， 功耗和速度比FP32
% posit 是xxx在数据格式，符合上述training的要求。具有xxx的特点
% 有人声明过其实用性，但是在training的应用还没有开发
% contribution :
%本文 对posit在low bit training中的应用做了系列的探索，
% 首次提出了关于posit的training的策略
% 利用低比特posit进行训练， 通过一些技巧来克服其精度问题， warm-up training , scale ,
% 之后实现了posit MAC单元， 可以用于日后的DNN accelerator
\section{Background}
\subsection{Reduced-Precision for DNN Training}
%当前的进展，一些training， 面临的问题
Training DNNs with reduced-precision is an appealing issue.
Gupta \textit{et al.} trained DNNs with fixed-point numbers, and introduced stochastic rounding procedure to prevent accuracy degradation\cite{gupta2015deep}.
% numerical limited
In paper\cite{miyashita2016convolutional}, the binary logarithmic data representation for both inference and training is explored, so that multiplication operations can be replaced by simpler shift operations.
% log
However, the above works usually can not provide expected model accuracy on complex tasks because there are too many information losses caused by the aggressive approximation.

% 精度不行
To deal with this problem, some recent works use reduced-precision floating point including FP8 or FP16 in training.
Micikevicius \textit{et al.}\cite{micikevicius2017mixed} used FP16 for forward and backward computation, and kept FP32 for weight update and accumulation. They also proposed a loss-scaling method to keep gradients propagation effectively.
% Mixed precision
Furthermore, with a chunk-based accumulation technique applied, Wang \textit{et al.}\cite{wang2018training} reduced the precision of the computation to FP8, and the precision of the weight update and accumulation to FP16.
% IBM FP-8
%By combining the floating point and fixed-point numbers,
%WAGE\cite{wu2018training} uses dynamic fixed-point format in training, but the layer-wise scaling factors need to be adjusted frequently to cover right dynamic ranges, which is a complicated process.

\subsection{Posit number system}
% 介绍
% posit 组成示意图
\begin{figure}[]
    \centering
    \includegraphics[width=0.48\textwidth]{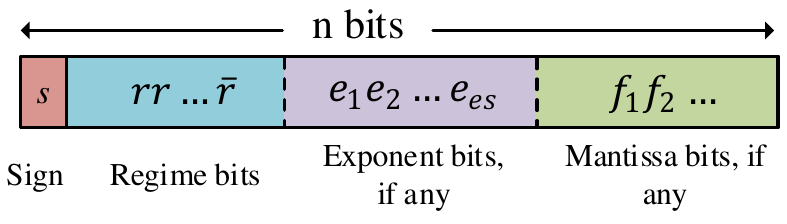}
    \caption{The basic structure of an $(n,es)$ posit number}
    \label{fig:posit_struct}
\end{figure}

% \begin{figure}[]
%     \centering
%     \includegraphics[width=0.48\textwidth]{regime.pdf}
%     \caption{The construction of $(5,1)$ posit number}
%     \label{fig:posit_ex}
% \end{figure}

 \begin{table}
     %\normalsize
     \centering
     \caption{The Detail Structures of Positive Values of $(5,1)$ Posit Number}
     \label{table:posit_ex}
     \begin{tabular}{|c|c|c|c|c|}
     \hline
      Binary Code &  Regime &  Exponent & Mantissa & Real Value\\
     \hline
     \textcolor{red}{0}\textcolor{reg}{0000}
             & \textcolor{reg}{x} & \textcolor{exp}{x}  & \textcolor{man}{x} & 0\\
     \textcolor{red}{0}\textcolor{reg}{0001}
             & \textcolor{reg}{-3} & \textcolor{exp}{0}  & \textcolor{man}{0} & 1/64\\
     \textcolor{red}{0}\textcolor{reg}{001}\textcolor{exp}{0}
             & \textcolor{reg}{-2} & \textcolor{exp}{0}  & \textcolor{man}{0} & 1/16\\
     \textcolor{red}{0}\textcolor{reg}{001}\textcolor{exp}{1}
             & \textcolor{reg}{-2} & \textcolor{exp}{1}  & \textcolor{man}{0} & 1/8\\
     \textcolor{red}{0}\textcolor{reg}{01}\textcolor{exp}{0}\textcolor{man}{0}                                                                          & \textcolor{reg}{-1} & \textcolor{exp}{0}  & \textcolor{man}{0} & 1/4\\
     \textcolor{red}{0}\textcolor{reg}{01}\textcolor{exp}{0}\textcolor{man}{1}                                                                          & \textcolor{reg}{-1} & \textcolor{exp}{0}  & \textcolor{man}{1/2} & 3/8\\
     \textcolor{red}{0}\textcolor{reg}{01}\textcolor{exp}{1}\textcolor{man}{0}
             & \textcolor{reg}{-1} & \textcolor{exp}{1}  & \textcolor{man}{0} & 1/2\\
     \textcolor{red}{0}\textcolor{reg}{01}\textcolor{exp}{1}\textcolor{man}{1}
             & \textcolor{reg}{-1} & \textcolor{exp}{1}  & \textcolor{man}{1/2} & 3/4\\
     \textcolor{red}{0}\textcolor{reg}{10}\textcolor{exp}{0}\textcolor{man}{0}
             & \textcolor{reg}{0} & \textcolor{exp}{0}  & \textcolor{man}{0} & 1\\
     \textcolor{red}{0}\textcolor{reg}{10}\textcolor{exp}{0}\textcolor{man}{1}
             & \textcolor{reg}{0} & \textcolor{exp}{0}  & \textcolor{man}{1/2} & 3/2\\
     \textcolor{red}{0}\textcolor{reg}{10}\textcolor{exp}{1}\textcolor{man}{0}
             & \textcolor{reg}{0} & \textcolor{exp}{1}  & \textcolor{man}{0} & 2\\
     \textcolor{red}{0}\textcolor{reg}{10}\textcolor{exp}{1}\textcolor{man}{1}
             & \textcolor{reg}{0} & \textcolor{exp}{1}  & \textcolor{man}{1/2} & 3\\
     \textcolor{red}{0}\textcolor{reg}{110}\textcolor{exp}{0}
             & \textcolor{reg}{1} & \textcolor{exp}{0}  & \textcolor{man}{0} & 4\\
     \textcolor{red}{0}\textcolor{reg}{110}\textcolor{exp}{1}
             & \textcolor{reg}{1} & \textcolor{exp}{1}  & \textcolor{man}{0} & 8\\
     \textcolor{red}{0}\textcolor{reg}{1110}
             & \textcolor{reg}{2} & \textcolor{exp}{0}  & \textcolor{man}{0} & 16\\
     \textcolor{red}{0}\textcolor{reg}{1111}
             & \textcolor{reg}{3} & \textcolor{exp}{0}  & \textcolor{man}{0} & 64\\

     \hline

     \end{tabular}
 \end{table}

An $(n, es)$ posit number, whose detail structure is shown as Fig. \ref{fig:posit_struct},
includes four parts: a sign bit, regime bits, $es$ exponent bits, and mantissa part.
The boundary between the last three parts are not fixed,
as the regime part is encoded by run-length method.
As for the numerical meaning of regime bits, consecutive $k~zeros$ ended by a $one$ means $-k$, consecutive $k+1~ones$ ended by a $zero$ means $k$.
As an example, a $(5,1)$ posit construction is described in Table \ref{table:posit_ex}.
% posit regime 编码形式
The value of a posit number $p$ (binary code) is given by Eq. (\ref{eq:posit}).

\begin{equation}\label{eq:posit}
    x =
    \begin{cases}
      0          & p = 000...0,  \\
      \pm\infty  & p = 100...0,  \\
      (-1)^s\times useed^k \times 2^e \times (1+f) & otherwise.
    \end{cases}
\end{equation}
where $useed=2^{2^{es}}$ determines the dynamic range.

The maximum and the minimum positive values that $p$ can represent are $useed^{n-2}$ and $useed^{2-n}$, respectively.
% posit 的特性

% posit 应用
Some groups have worked on the design of hardware architecture generators for posit arithmetics.
Jaiswal \textit{et al.}\cite{jaiswal2018universal} proposed a parameterized posit arithmetic architectures generator,
supporting basic operations such as FP-Posit conversion, addition/subtraction, and multiplication.
Recently, an efficient posit MAC unit generator that can be combined with a reasonable pipeline strategy
was put forward by Zhang \textit{et al.}\cite{zhang2019efficient},
% and a reasonable pipeline strategy is also introduced.
Besides, the applications of low-bit posit in deep learning also attracted some attentions.
Deep Positron\cite{carmichael2018deep}, a DNN architecture that employs exact-multiply-and-accumulates (EMACs) for 8-bit posit,
shows better accuracies than 8-bit fixed-point and FP for some small datasets.
J.Johnson\cite{johnson2018rethinking} proposed log-float format inspired by posit, and use it for DNN inference,
whose accuracy loss is less than $1\%$ for ImageNet dataset within ResNet-50 model.
% 应用

\section{Posit Training Strategy and Experiments Results}

\begin{table}
    \centering
    \caption{Notations For Posit Transformation }
    \label{table:notation}
    \begin{tabular}{l|l}
    \hline
     Name &  Description \\
    \hline
        $n$     &     posit word size      \\
        $es$    &     posit exponent field size  \\
        $s$     &     sign of the number  $x$\\
        $exp$   &     the effective exponent value of $x$ \\
        $k$     &     the regime value of $px$\\
        $e$     &     the exponent value of $px$ before rounding\\
        $f$     &     the mantissa value of $px$ before rounding \\
        $kb$    &     the regime width of $px$ \\
        $eb$    &     the exponent width of $px$ \\
        $fb$    &     the mantissa width of $px$ \\
        $pe$    &     the exponent value of $px$ after rounding \\
        $pf$    &     the mantissa value of $px$ after rounding \\
    \hline

    \end{tabular}
\end{table}

\subsection{Posit Transformation}
% 介绍 float 到posit的转换过程
% 先说为甚要介绍：如何将一个数转换为posit 表示，
In this work, all data and computations are represented in posit format in the training process.
Therefore, we have to transform a real number, which is represented in FP32 format in current computers, to posit format. Here we define an operator $P_{n,es}(x)$ to achieve this task. The detail process is shown in Algorithm \ref{alg:transform}, and the involved notations are listed in Table \ref{table:notation}.

% 简要介绍转换过程
Given the total word size $n$ and exponent field size $es$, we can determine the dynamic range of a posit number.
To convert a non-zero number $x$ to corresponding posit number $px$,
firstly we have to limit its magnitude based on the dynamic range and
then extract sign, regime, exponent, and mantissa parts.

\begin{algorithm}\label{alg:transform}
    \caption{Transform a Number to Posit Format}
    \KwIn{real number $x$, posit word size $n$ and exponent field size $es$}
    \KwOut{posit number $px$ }
    $useed = 2^{2^{es}}$ \;
    $maxpos = useed^{n-2}, ~ minpos = useed^{2-n}$\;
    \SetAlgoVlined
    \uIf{$abs(x)  < minpos$}
    {
        $px = 0$ \;
    }
    \Else
    {
        $s = sign(x)$ \;
        $x' = clip\{abs(x), minpos, maxpos\} $\;
        $exp = \lfloor log_2{x'} \rfloor$ \;
        $k = \lfloor exp\div 2^{es} \rfloor$ \;
        $e = exp - k\times 2^{es}$ \;
        $f = x' / 2^{exp} - 1 $ \;
        \uIf{$k \ge 0$}
            {
                $rb = k + 2 $ \;
            }
        \Else{
                $rb = -k + 1 $ \;
            }
        $eb = min\{n-1-rb, es\}$ \;
        $fb = min\{n-1-rb-eb, 0\} $ \;
        $pe = \lfloor e\times 2^{eb-es} \rfloor \times 2^{es-eb}$\;
        $pf = \lfloor f \times 2^{fb} \rfloor \times 2^{-fb} $\;

        $px = s\times useed^{k} \times 2^{pe} \times ( 1 + pf) $ \;
    }

    \Return {$px$} \;
\end{algorithm}

\begin{figure}[H] %这里使用的是强制位置，除非真的放不下，不然就是写在哪里图就放在哪里，不会乱动
	\centering  %图片全局居中
	\vspace{-0.35cm} %设置与上面正文的距离
	\subfigtopskip=2pt %设置子图与上面正文或别的内容的距离
	\subfigbottomskip=2pt %设置第二行子图与第一行子图的距离，即下面的头与上面的脚的距离
	\subfigcapskip=-5pt %设置子图与子标题之间的距离
	\subfigure[histgram of $conv1.weight$]{
		\label{conv_hist}
		\includegraphics[width=0.45\linewidth]{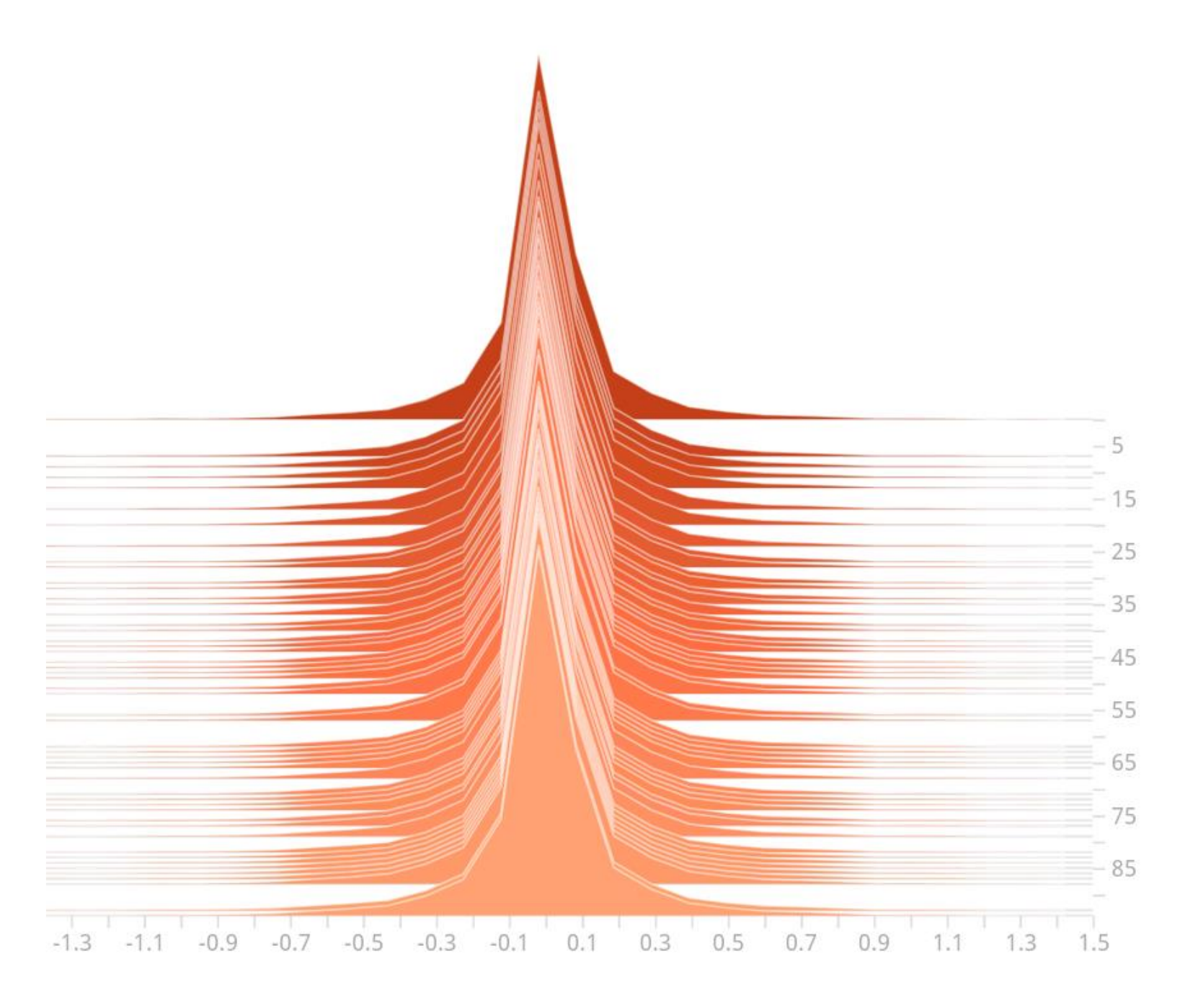}}
	%\quad %默认情况下两个子图之间空的较少，使用这个命令加大宽度
	\subfigure[distribution of $conv1.weight$]{
		\label{conv_dist}
		\includegraphics[width=0.45\linewidth]{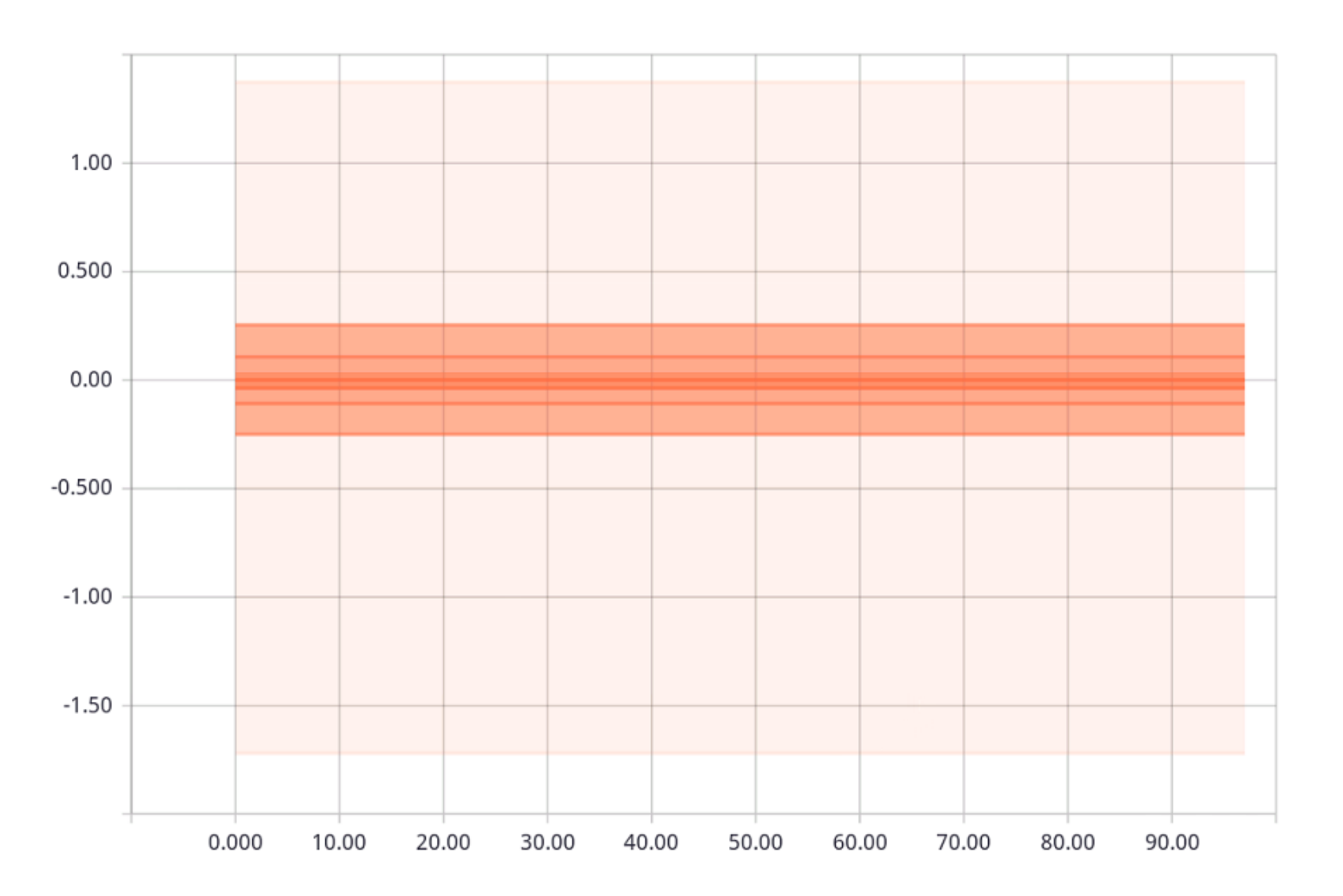}}
      %这里是空了一行，能够实现强制将四张图分成两行两列显示，而不是放不下图了再换行，使用\\也行。

	\subfigure[histgram of $bn4.0.1.weight$]{
		\label{bn_hist}
		\includegraphics[width=0.45\linewidth]{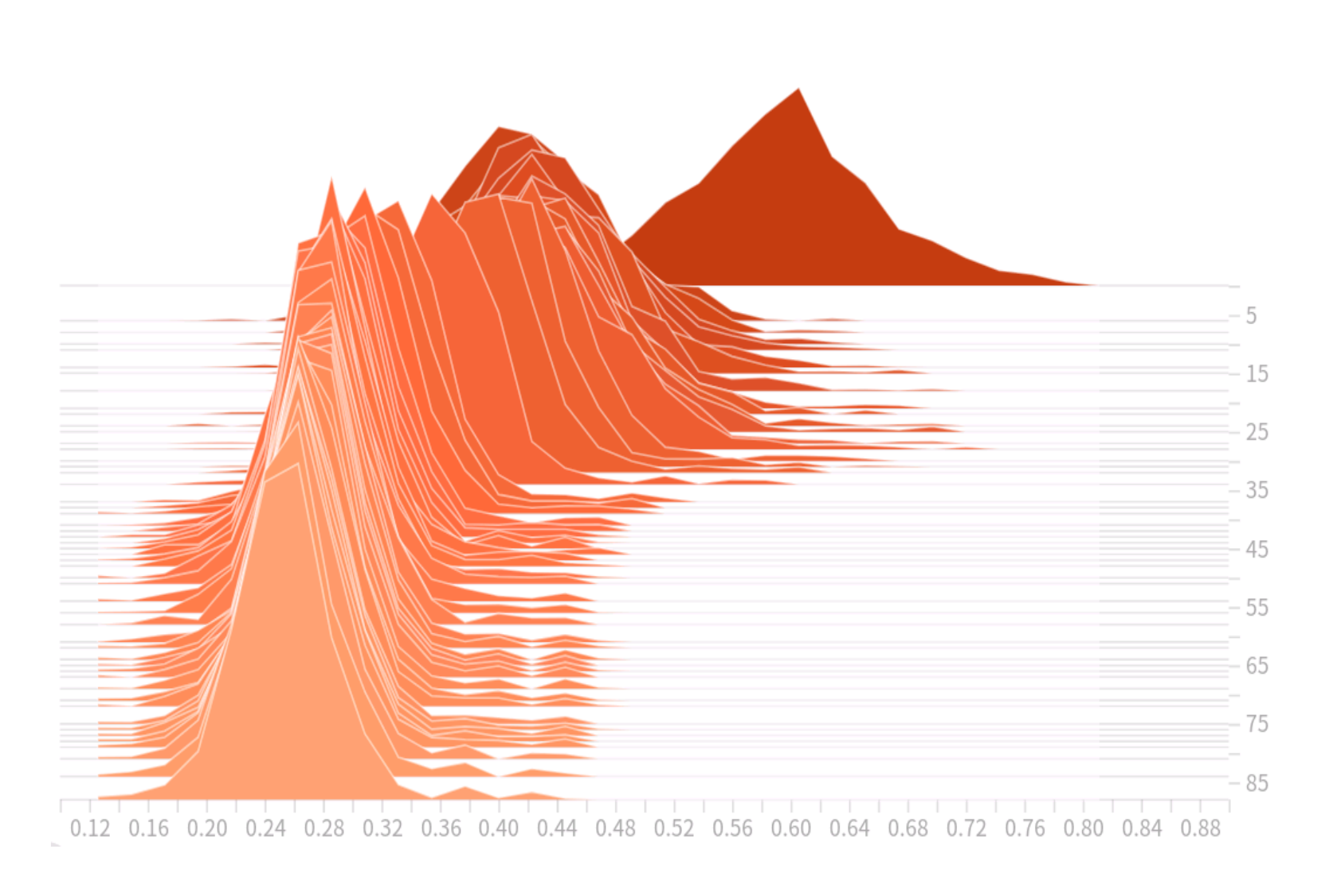}}
	%\quad
	\subfigure[distribution of $bn4.0.1.weight$]{
		\label{bn_dist}
		\includegraphics[width=0.45\linewidth]{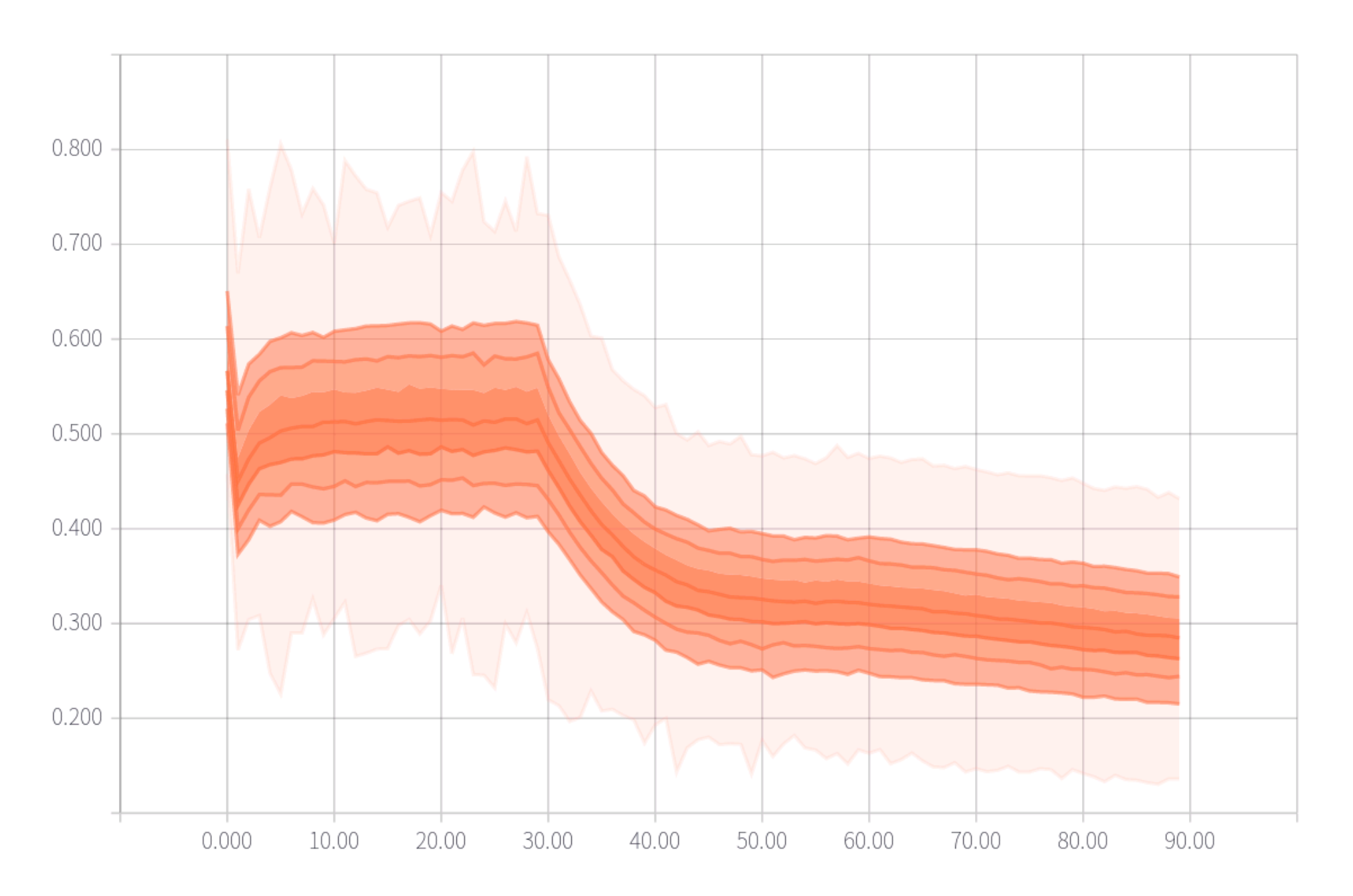}}
	\caption{The histgrams and distributions of CONV layer and BN layer in training process}
	\label{fig:hist_dist}
\end{figure}

\begin{figure*}[!htp]
    \subfigure[Forward propagation with posit transformation]{
        \label{state1}
        \includegraphics[scale=0.5]{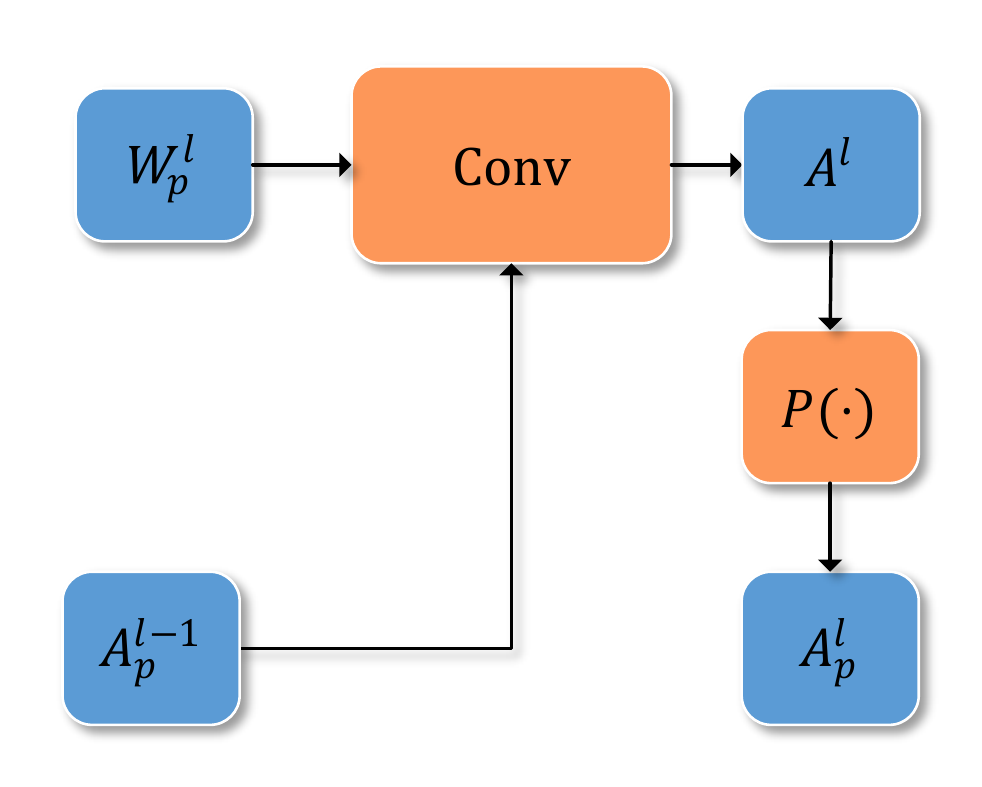}}
    \hspace{0.1in}
    \subfigure[Backward propagation with posit transformation]{
        \label{state2}
        \includegraphics[scale=0.5]{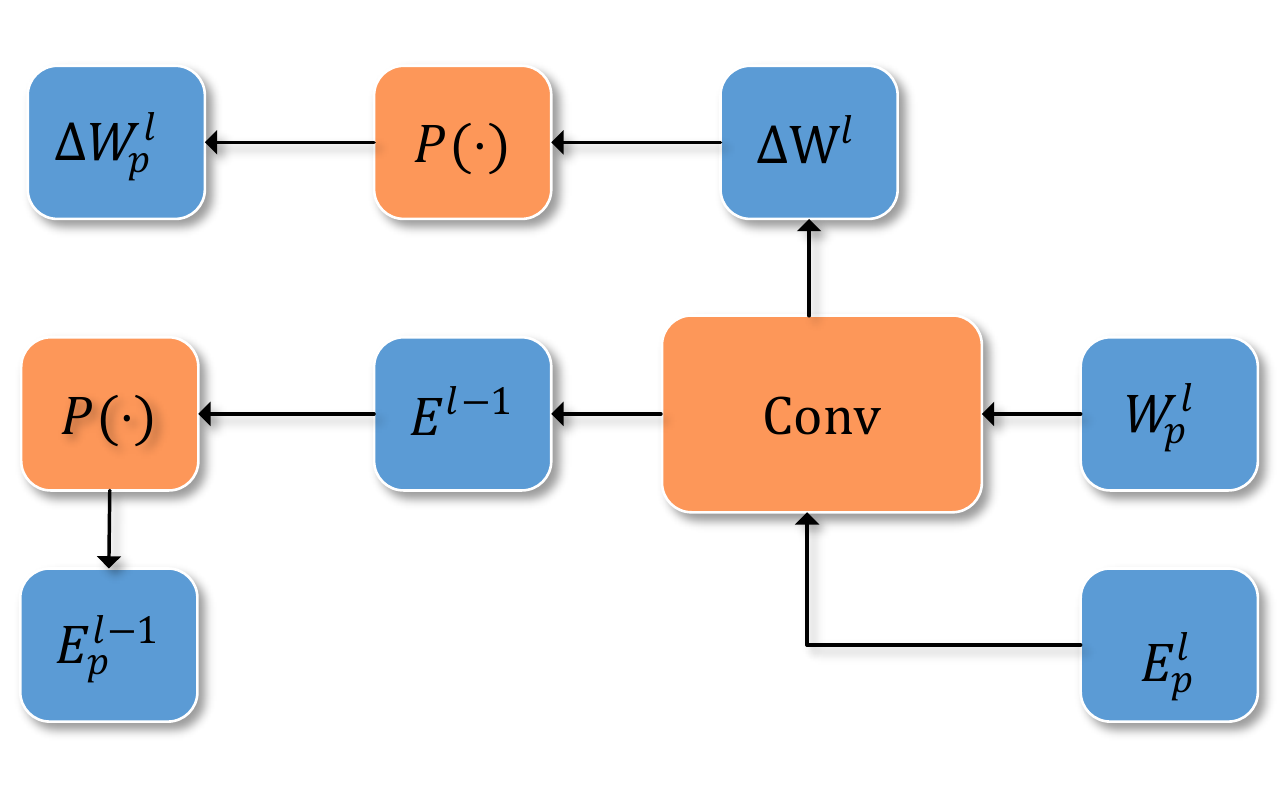}}
    \hspace{0.1in}
    \subfigure[Weight update with posit transformation]{
        \label{state3}
        \includegraphics[scale=0.5]{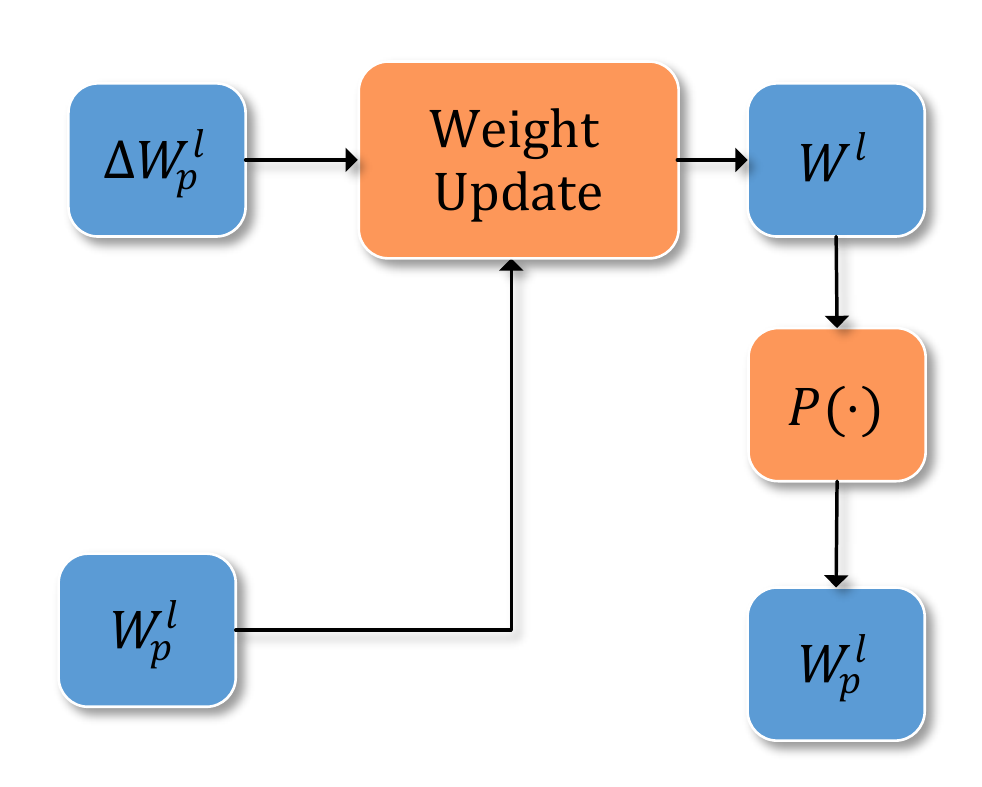}}
    \caption{DNN training computation flow graph with posit transformation.
    In the graph, $P(\cdot)$ means the transformation operation, whose subscript $(n,es)$ is omitted for simplicity.
    Besides, $W$, $A$, $\Delta W$, and $E$ stand for the weight, activation, weight gradient, and error in a layer respectively.
    The symbols with subscript $p$ are in posit format.
     }
    \label{fig:insert_flow}
\end{figure*}

% 将整个训练策略整合
% 1. 考虑到分布问题， 希望能够充分利用 编码效率
% 2. 观察到分布 在巡礼过程中是基本稳定的，因此采用 合适的scale
% 3. 观察到posit的精度分布，和
% 4. 考虑到posit精度和 数据分布 能够match， 也就是分布密集的部分具有 更高的精度 (说的过去嘛？)

Next, because of the restriction of word size, the width of each part is adjusted.
Therefore, the rounding operations are applied to the value of each part to fit the
adjusted width.
Here we choose the rounding-to-zero method, e.g. the $\lfloor \cdot \rfloor$ operator in Algorithm \ref{alg:transform}, Line 16, 17.
Comparing to the rounding-to-nearest and stochastic rounding methods,
the rounding-to-zero will be more friendly for hardware implementation.
Finally, the posit result $px$ is attained by combining these parts based on Eq. (\ref{eq:posit}).
% 没有用 stochastic rounding

With the transformation algorithm accomplished, we insert it in DNN training computation flow as depicted in Fig. \ref{fig:insert_flow},
which includes forward process, backward process, and weight update process.

\subsection{Training a DNN Model with Posit}

Although posit has many benefits while being used in DNN training, it can not show expected performance if we replace FP32 with reduced-precision posit directly.
There are several key reasons as follows:
    \begin{itemize}
        \item In the early training stage, the model is more sensitive to the precision of data,
        and the distributions of some layers are unstable,
        % !!!!!!!!!!!!!!!!
        so that the reduced-precision representations will cause a bad initialization and make the model hard to converge.
        \item In fact, the precision of posit number system is basically symmetrical about 1, but the data distributions in DNN models are concentrated on limited range. To some extent, it results mismatching between data distributions and number representation formats, thereby leading larger approximation errors.
        % It results that they do not match well to some extent
        \item For different layers, the data have different ranges, which means some data distributions are more concentrated and the others are relatively decentralized. Therefore, it is sub-optimal to use same data precision (e.g $es$ of posit) to represent them.
    \end{itemize}
In this section, we propose corresponding methods for dealing with the above problems.

\textbf{Warm-up Training:}$~$
By observing the distributions of data in training process, we find that most of them are approximately normal.
As shown in Fig. \ref{fig:hist_dist}, the distributions of the weights in Convolution (CONV) layers are basically stable in the training process.
However, because of the initialization method, the distributions of the weights in Batch Normalization (BN) layers have a steep change in the first several epochs, which may be an important reason of high model sensitivity in early training process.
Therefore, in this phase, a higher numerical precision is required.
% 不好啊!!!!!!!!!
On account of this situation, a warm-up training using FP32 for several epochs (1-5 epochs) is carried out.
It will be helpful to determine the data distribution effectively and make sure the convergence of networks.

\textbf{Distribution-based Shifting:}$~$
When transforming a real number $x$ to its reduced-precision format,
the most common idea is approximating it to the nearest reduced-precision value
and clipping it based on the dynamic range of reduced-precision format.
As a result, the numerical errors are inevitable.
To overcome the second issue, a scaling factor is introduced to shift the data distribution to a more appropriate range,
whose upper bound is usually the maximum value that the reduced-precision number can represent \cite{wu2018training}.
As for posit number system, its dynamic range is large enough to meet demand.
However, to make full use of the code space of posit, inspired by the shift-based mapping method \cite{wu2018training},
we also propose a layer-wise scaling factor $S_f$.
The calculation of the scaling factor is shown as Eq. (\ref{eq:sf}).
\begin{equation}\label{eq:sf}
    \begin{aligned}
        center  &= round(mean(\log_2(x))), \\
        S_f       &= 2^{(center + \sigma)}. \\
    \end{aligned}
\end{equation}
$x$ is a tensor to be converted, $center$ means the approximate distribution center of the input tensor in log domain,
which stands for that the majority of values are close to this magnitude,   %什么啊
$\sigma$ is a predefined positive integer constant, which is set as 2 in our experiments.
As mentioned in previous works\cite{han2015deep}, the large values have more importance than small values,
so we add $\sigma$ to $center$ for shifting values towards small magnitude a little more.
Basd on the warm-up trained model, the scaling factor of each layer can be calculated.
Finally, by applying the scaling factor before and after transformation operation $P(x)$ as Eq. (\ref{eq:asf}),
the more important values are shifted to the order of magnitude that has higher precision.

\begin{equation}\label{eq:asf}
    px = P(\frac{x}{S_f})S_f.
\end{equation}

\textbf{Adjust Dynamic Range:}$~$
During the DNN training process, different layers have different distribution ranges which are measured approximately by the difference between
the maximum and minimum value in log domain.
For example, in the first few layers, the ranges of gradients are relatively larger than the ranges of other values.
In this case, the posit number should have a larger dynamic range, which means a bigger $es$ value.
In this work, for simplicity, we just set the $es$ to be 1 for all weights and activations, and be 2 for all gradients and errors.

\subsection{Experiment Results}

To validate our posit training strategy, we perform experiments with ResNet-18\cite{he2016deep} on ImageNet and Cifar-10 datasets utilizing Pytorch framework on NVIDIA P100 GPUs.
The validate top-1 accuracy and related configuration are summarized in Table \ref{table:al_results}. which demonstrate that training with reduced-precision posit number can achieve FP32 baseline accuracy without tuning hyperparameters.
The training details are as follows:

    \textbf{Cifar-10:}$~$The model uses stochastic gradient descent with moment 0.9 as optimizer. The initial learning rate is set to 0.1 and  divided by 10 at epoch 60, epoch 150, and 250.
    The network is trained for 300 epochs with a mini-batch size of 512.
    The warm-up training runs for 1 epoch.
    %As for data augmentation, we follow the method in \cite{lee2014deeply}.

    \textbf{ImageNet:}$~$The model uses stochastic gradient descent with moment 0.9 as optimizer. The initial learning rate is set to 0.1 and  divided by 10 every 30 epochs.
    The model is trained for 90 epochs with a mini-batch size of 512.
    The warm-up training runs for 5 epochs.

\begin{table}
    \centering
    \caption{Training Configurations and Validate Accuracies Results}
    \label{table:al_results}
    \begin{threeparttable}
    \begin{tabular}{l|l|l}
    \hline
    \hline
        Dataset &   Cifar-10    &   ImageNet    \\
        %batch size & epochs & posit             & posit (acc1) & FP32 baseline  \\
    \hline
        model                   &      Cifar-ResNet-18  &  ResNet-18 \\
        batch size              &      512              &  512              \\
        epochs                  &    300                &   120              \\
        optimizer               &    SGD with Moment    & SGD with Moment \\
    \hline
    \hline
        FP32 baseline             &    93.40            &   71.02           \\
        posit                   &    92.87$^{\mathrm{1}}$ & 71.09$^{\mathrm{2}}$ \\
       % posit-2$^{\mathrm{2}}$    &      -              &   66.64           \\
        %posit-3$^{\mathrm{3}}$    &    92.87            &    65.30          \\
    \hline
    \end{tabular}
    \begin{tablenotes}
        \footnotesize
        \item[1] posit (8,1) for CONV layers forward pass and weight update, posit (8,2) for CONV layers backward pass.
        posit (16,1) for BN layers forward pass and weight update, posit (16,2) for BN layers backward pass.
        \item[2] posit (16,1) for forward pass and weight update, posit (16,2) for backward pass.

        %\item[2] posit (8,1) for forward pass and weight update, posit (16,2) for backward pass.
        %\item[3] posit-3: (8,1) for Conv layers forward pass, (8,2) for gradient and error.
    \end{tablenotes}
    \end{threeparttable}
\end{table}

\section{Energy-Efficient Posit MAC Architecture}

% 训练过程中 时间和能耗都是一见非常令人关心的事。
% 在DNN 加速器中 这两项overhead 主要来自于 大量重复的 乘累加计算， 和多次大量的的数据访问过程。
%　８－１６位的posit 能够将存储的数据量降低为 2-4倍，对数据访问的开销能够大大降低
% 对于主要的计算过程， 如果想要充分利用起 posit的潜在优势，那么一个支持posit计算的MAC单元是必要的

% 当前关于posit单元的设计，多数是convert->float->convert的思路。一个设计不好的convert单元会对延时造成很坏的影响。
%
By using 8 bits or 16 bits posit number for training, the model size can be reduced to 25\% or 50\%, then the energy consumption can be saved significantly, because the memory space requirements and the communication bandwidth are reduced.
As for computational process, the energy consumption mainly comes from a mass of MAC operations. Since the posit arithmetic operations are different from traditional floating point arithmetic operations, a dedicated MAC unit is urgently required to take full advantage of the reduced-precision posit.

\begin{figure}[]
    \centering
    \includegraphics[width=0.35\textwidth]{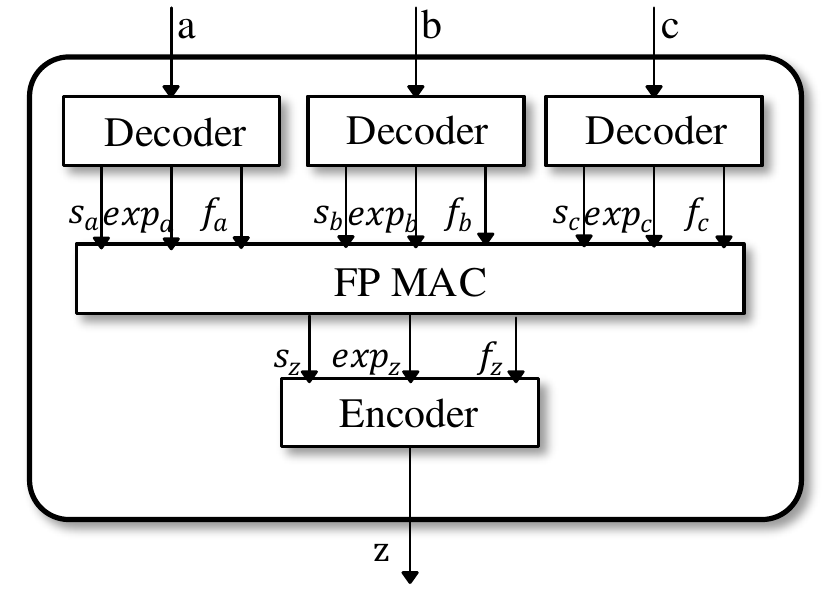}
    \caption{The overall architecture for the posit MAC}
    \label{fig:posit_mac}
\end{figure}

As shown in Fig. \ref{fig:posit_mac}, the posit MAC unit proposed in \cite{zhang2019efficient} mainly compose of three units: a decoder converting posit to FP, an FP MAC unit, and an encoder converting FP to posit. In this way, the summation of the encoder delay and decoder delay consumes about 40\% time of the total posit MAC delay.

\begin{figure}[H]
	\centering
	\vspace{-0.35cm}
	\subfigtopskip=2pt
	\subfigbottomskip=2pt
	\subfigcapskip=-5pt
	\subfigure[The original encoder]{
		\label{original_encoder}
		\includegraphics[width=0.48\linewidth]{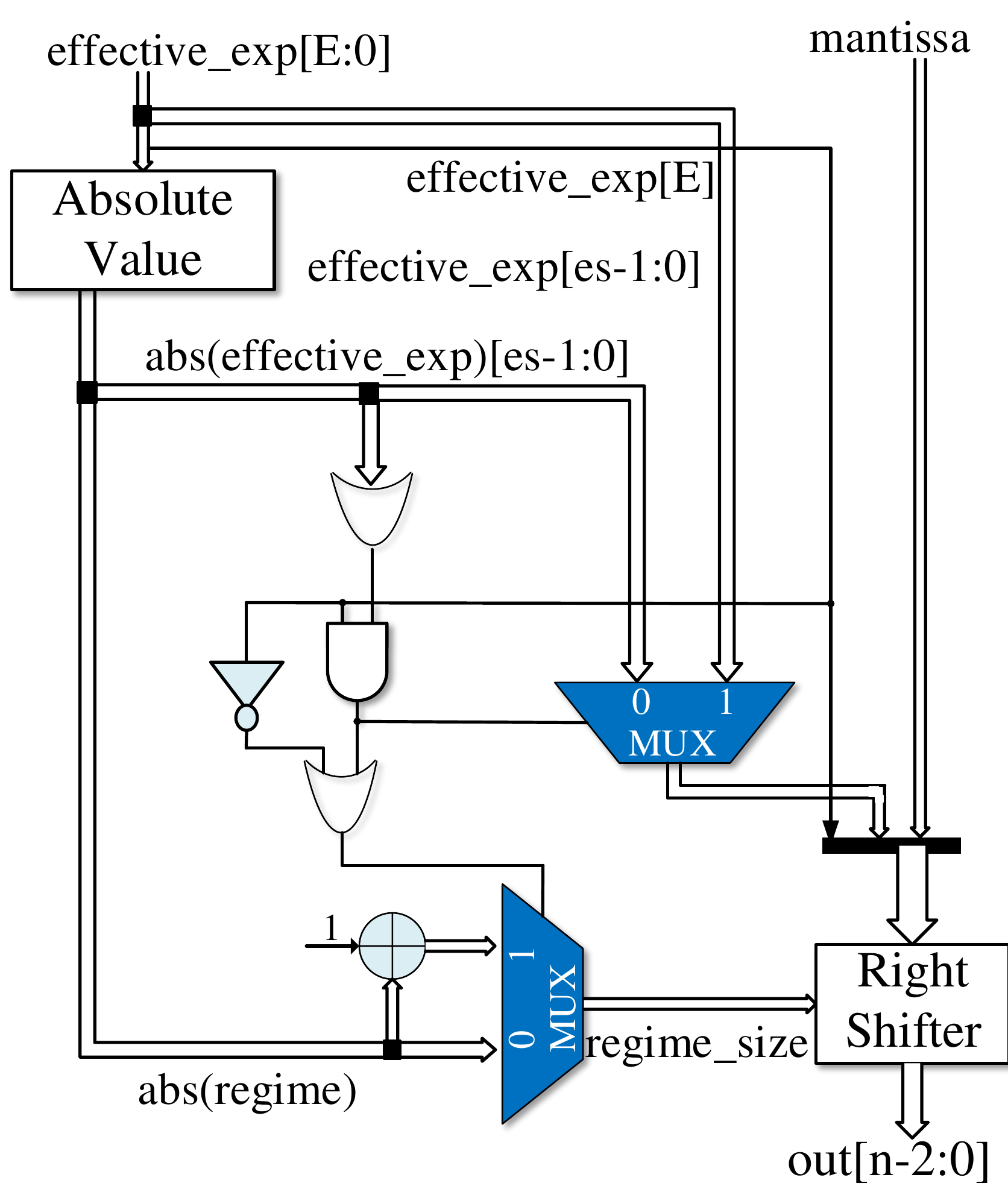}}
	%\quad %默认情况下两个子图之间空的较少，使用这个命令加大宽度
	\subfigure[The optimized encoder]{
		\label{optimized encoder}
		\includegraphics[width=0.48\linewidth]{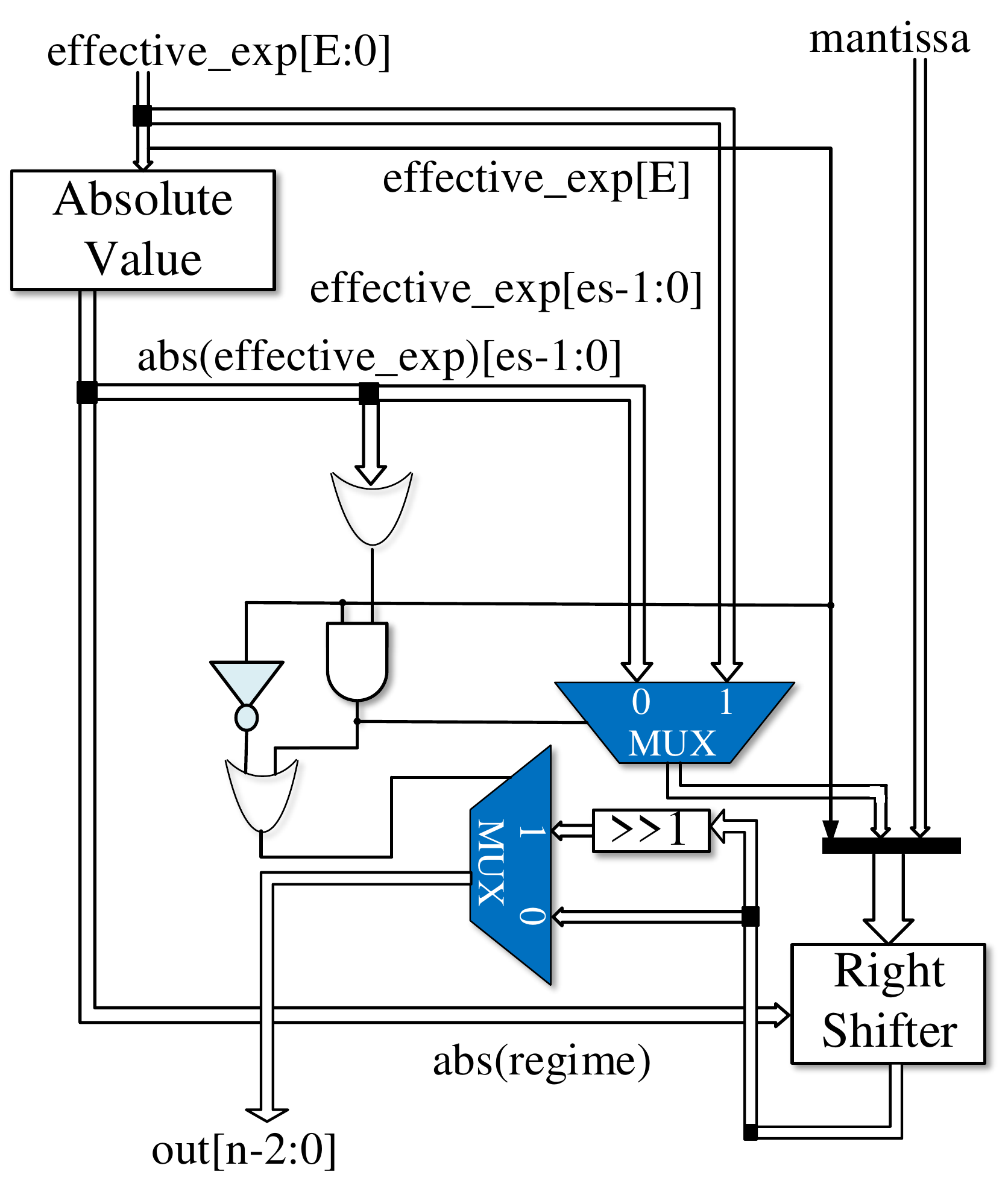}}
      %这里是空了一行，能够实现强制将四张图分成两行两列显示，而不是放不下图了再换行，使用\\也行。
	\caption{The encoder architectures before and after optimization}
	\label{fig:encoder}
\end{figure}

Based on this result, improved architectures for the encoder and decoder with lower latency are proposed, which are shown in
Fig. \ref{fig:encoder} and Fig. \ref{fig:decoder}.

\subsection{The Optimized Decoder and Encoder Architectures}
\begin{figure}[H]
	\centering
	\vspace{-0.35cm}
	\subfigtopskip=2pt
	\subfigbottomskip=2pt
	\subfigcapskip=-5pt
	\subfigure[The original decoder]{
		\label{original_decoder}
		\includegraphics[width=0.48\linewidth]{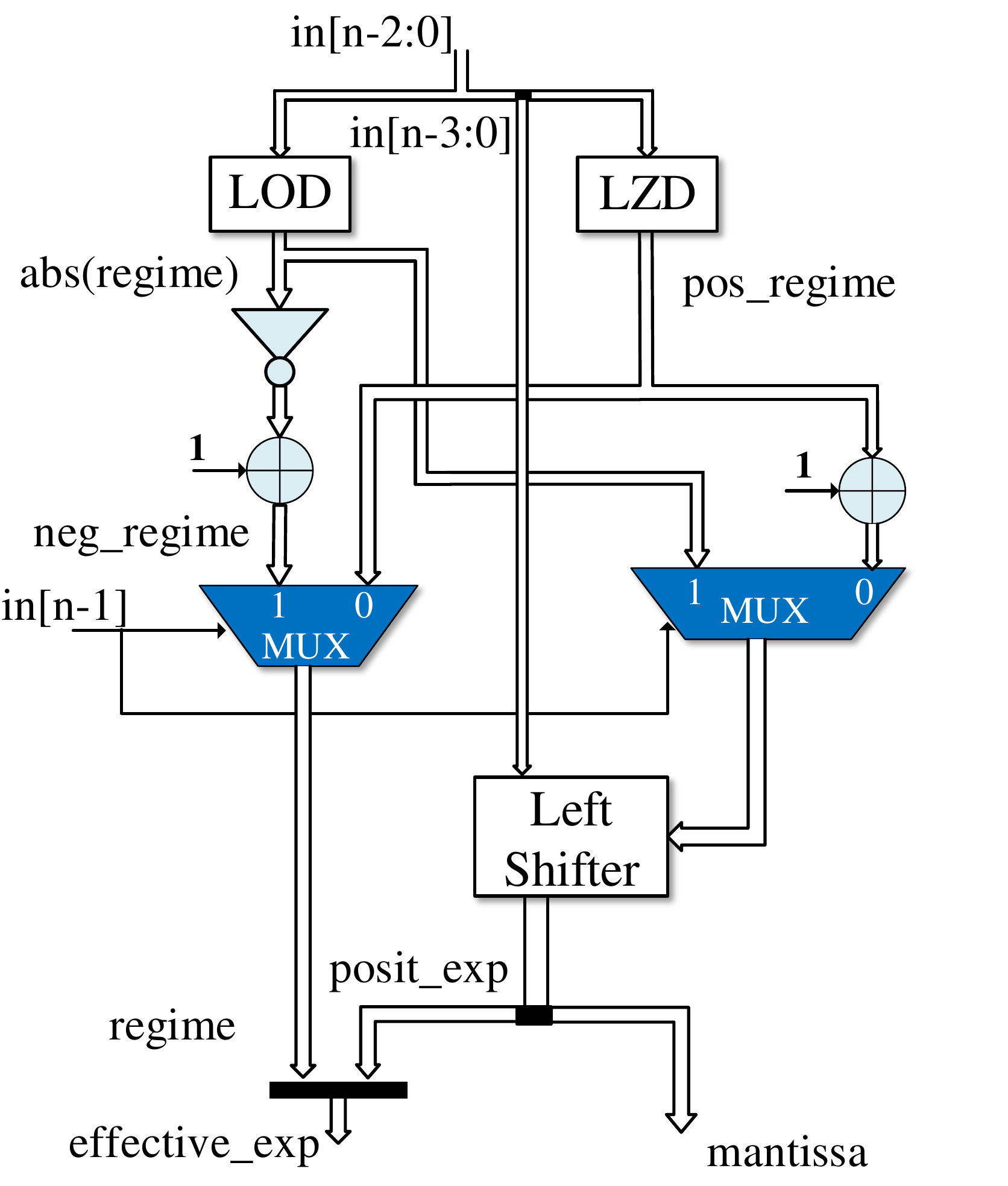}}
	%\quad %默认情况下两个子图之间空的较少，使用这个命令加大宽度
	\subfigure[The optimized decoder]{
		\label{optimized decoder}
		\includegraphics[width=0.48\linewidth]{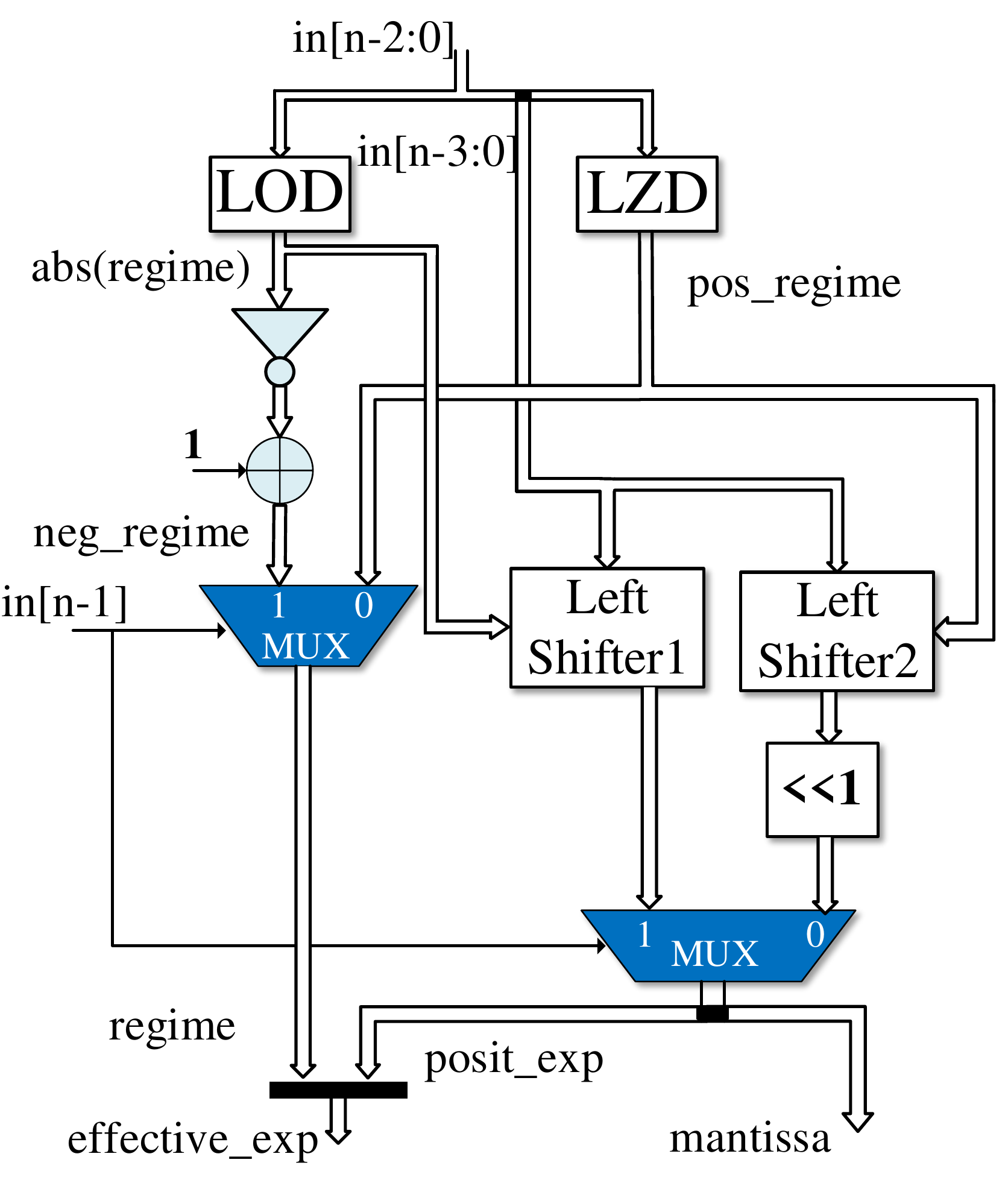}}
      %这里是空了一行，能够实现强制将四张图分成两行两列显示，而不是放不下图了再换行，使用\\也行。
	\caption{The decoder architectures before and after optimization}
	\label{fig:decoder}
\end{figure}

The decoder aims to extract different parts of posit, then exports effective exponent value and mantissa value.
Firstly, the absolute regime value of the input posit number is calculated by a LOD (if real regime value is negative) or a LZD (if real regime value is positive).
Secondly, The input is left shifted by the width of regime bits, which is equal to $r$ or $r+1$, where $r$ is the absolute regime value.
The output of $Left~Shifter$ composes of posit exponent value and mantissa value.
Finally the regime value and posit exponent value are packaged into effective exponent value.
The critical path of the original decoder is determined by the add one operation.
As shown in Fig. \ref{fig:decoder}, we remove the adder, and split the left shift path by duplicating the $Left~Shifter$.
To preserve the function of the adder, a left-shift-one ($``<<1''$ ) operation is inserted after the $Left~Shifter2$.

% The original encoder\cite{jaiswal2018universal} and optimized encoder are
The encoder converts the FP to posit format.
Firstly, a 2n-bit variable $REM$ is constructed with mantissa and the least significant bits(LSB) $es$ exponent bits, and the remained bits are filled by regime sequence.
Then $REM$ is right shifted by the width of regime bits, which is equal to $r$ or $r+1$, where $r$ is the absolute regime value.
Therefore, an optimization method, which is similar to that used in the optimized decoder, is applied for the encoder architecture.

\subsection{Hardware Implementation Results}
% 首先说 实现环境， 实现目标
The architectures are coded by Verilog HDL and synthesized by Design Compiler under TSMC 28nm technology.
To prove efficiency of the proposed encoder and decoder, the same parameterized architectures with \cite{zhang2019efficient}
are evaluated.
\begin{table}
    \centering
    \caption{Delay Comparison of Encoder and Decoder with \cite{zhang2019efficient}}
    \label{table:encoder_decoder}
    \begin{threeparttable}
    \begin{tabular}{l|l|l|ccc}

    \hline
    \multicolumn{3}{c|}{}                        &   posit(8,0)     &   posit(16,1)   & posit(32,3) \\

    \hline
    \multirow{2}{*}{\cite{zhang2019efficient}}  &
        \multirow{2}{*}{delay(ns)}  &  encoder  &  0.2             & 0.29            & 0.35         \\
          &                          &  decoder  &  0.2             & 0.28            & 0.34         \\

    \hline
    \multirow{6}{*}{Ours}
        &   \multirow{2}{*}{delay(ns)}  &  encoder  &  0.13             & 0.18            & 0.23         \\
        &                          &  decoder  &  0.14             & 0.21            & 0.29         \\
        \cline{2-6}
        &   \multirow{2}{*}{power(mW)}  &  encoder  &  0.21             & 0.44            & 0.59        \\
        &                          &  decoder  &  0.27             & 0.45            & 0.66         \\
        \cline{2-6}
        &   \multirow{2}{*}{area($\mu m^2$)}    &  encoder  &  137              & 295               & 540         \\
        &                                       &  decoder  &  201              & 504               & 960        \\

    \hline
    \end{tabular}
    \end{threeparttable}
\end{table}

The comparison results in Table \ref{table:encoder_decoder} show our encoder speeds up by 25\%-35\% and our decoder speeds up by 15\%-30\%, thereby reducing the impact of these two units on total delay.

After combining the proposed encoder and decoder with the FP MAC unit, an energy-efficient posit MAC architecture is proposed.
To meet the requirements of the DNN training with posit, different posit MAC units which support all kinds of posit format involved in Table \ref{table:al_results} are implemented.
The implementation results are summarized in Table \ref{table:hw_results}.
For fair comparison between the posit MAC and FP32 MAC on energy consumption, all these units are synthesized with a timing constraint of 750MHz.
Comparing to FP32 MAC, the posit MAC can reduce the power by 22\%-83\%, and reduce the area by 6\%-76\%.

\begin{table}
    \centering
    \caption{Comparison of Posit MAC with FP32}
    \label{table:hw_results}
    \begin{threeparttable}
    \begin{tabular}{l|l|l}

    \hline
                            &  Power(mW)     &   Area ($\mu m^2$)    \\
    \hline
        FP32                &    2.52          &  4322              \\
        posit(8,1)          &    0.45          &  1208              \\
        posit(8,2)          &    0.35          &  1032              \\
        posit(16,1)         &    1.77          &  4079              \\
        posit(16,2)         &    1.60          &  3897              \\

    \hline
    \end{tabular}
    \end{threeparttable}
\end{table}

\section{Conclusion and Future Work}
In this paper, with several useful methods proposed, the posit number system is applied to DNN training successfully.
The experiments results show that reduced-precision posit can achieve similar accuracy with FP32 on different datasets.
If the posit is applied in DNN accelerators, the overhead caused by data communications can be saved by 2-4$\times$.
In order to take full advantage of posit, an energy-efficient posit MAC unit is designed.
Comparing to FP32 MAC, the posit MAC can reduce the power by 22\%-83\%, and reduce the area by 6\%-76\%.

In the further work, we will implement a hardware accelerator for DNN training with posit.
On the other hand, the architectures for posit arithmetic with the encoder and decoder may be not the optimal method.
%It still needs to be exp to find a more suitable method.
We will carefully design a new architecture for the posit MAC to further improve its performance.

\bibliographystyle{unsrt}
\bibliography{bibitem}

\end{document}